\documentclass[10pt]{wlscirep}
\usepackage[utf8]{inputenc}
\usepackage[T1]{fontenc}
\usepackage{tikz}
\usetikzlibrary{shapes.geometric, arrows.meta, positioning}
\usepackage{makecell}
\usepackage{tabularx}
\usepackage{array}
\usepackage{multirow}
\usepackage{placeins}
\usepackage{subcaption}
\usepackage[cal=cm]{mathalfa}
\usepackage{caption} 
\usepackage{subcaption}
\hypersetup{hidelinks}
\usepackage{xr}
\title{A multicenter benchmark and clinically structured metric for coronary CTA report generation}

\author[1,2,4,+]{Zhiyu Ye}
\author[3,+]{Yue Sun}
\author[3]{Limiao Zou}
\author[3]{Cheng Xu}
\author[3]{Keting Xu}
\author[2]{Tong Hu}
\author[2]{Yue Yu}
\author[1,*]{Hairong Zheng}
\author[4,*]{Yining Wang}
\author[2,*]{Tong Zhang}

\affil[1]{State Key Laboratory of Biomedical Imaging Science and System, Shenzhen Institute of Advanced Technology, Chinese Academy of Sciences, Shenzhen, China}
\affil[2]{Pengcheng Laboratory, Shenzhen, China}
\affil[3]{Department of Radiology, State Key Laboratory of Complex Severe and Rare Diseases, Peking Union Medical College Hospital, Chinese Academy of Medical Sciences and Peking Union Medical College, Beijing, China}
\affil[4]{University of Chinese Academy of Sciences, Beijing, China}

\affil[*]{Correspondence: zhangt02@pcl.ac.cn}
\affil[+]{These authors contributed equally to this work.}


\begin{abstract}
Reliable evaluation of automated coronary computed tomography angiography (CCTA) report generation requires standardized multicentre benchmarks and clinically structured metrics. We established a four-centre benchmark comprising 3,021 CCTA series from 818 patient-report pairs to evaluate seven open-source three-dimensional vision-language models.
We developed CSM$_{\text{CCTA}}$, a clinically structured metric for CCTA report evaluation, with patient-, vessel-, and segment-level variables defined according to clinical guidelines. Report pairs are compared at the finest shared anatomical level, and the contributions of different clinical components are weighted based on expert assessments. We estimated these weights using 70 expert-scored cases and evaluated clinical alignment in a non-overlapping set of 30 cases. CSM$_{\text{CCTA}}$ showed a strong correlation with radiologist scores (Pearson's $r=0.97$, $p<0.001$), exceeding the next-best metric, FORTE ($r=0.70$), by 0.27, and agreed with expert preferences in 115 of 160 pairwise comparisons (71.9\%). Under controlled perturbations, CSM$_{\text{CCTA}}$ remained stable to clinically equivalent wording and decreased monotonically with progressive information omission. In the multicenter benchmark, the CCTA-trained C2RG model achieved the highest CSM$_{\text{CCTA}}$ scores across all four hospitals, although its performance remained far from optimal. In contrast, CCTA-irrelevant reports accounted for up to 98.7\% of the outputs from generalist models. Together, the benchmark provides a standardized setting for model comparison, while CSM$_{\text{CCTA}}$ enables clinically structured evaluation of finding agreement and anatomical specificity. These results support a more clinically aligned and anatomically resolved approach to evaluating CCTA report generation. Code is available at \url{https://openi.pcl.ac.cn/OpenMedIA/CSM_CCTA}.
\end{abstract}
\begin{document}

\flushbottom
\maketitle
%
%
\thispagestyle{empty}


\begin{figure}[!htbp]
\centering
\includegraphics[width=\linewidth]{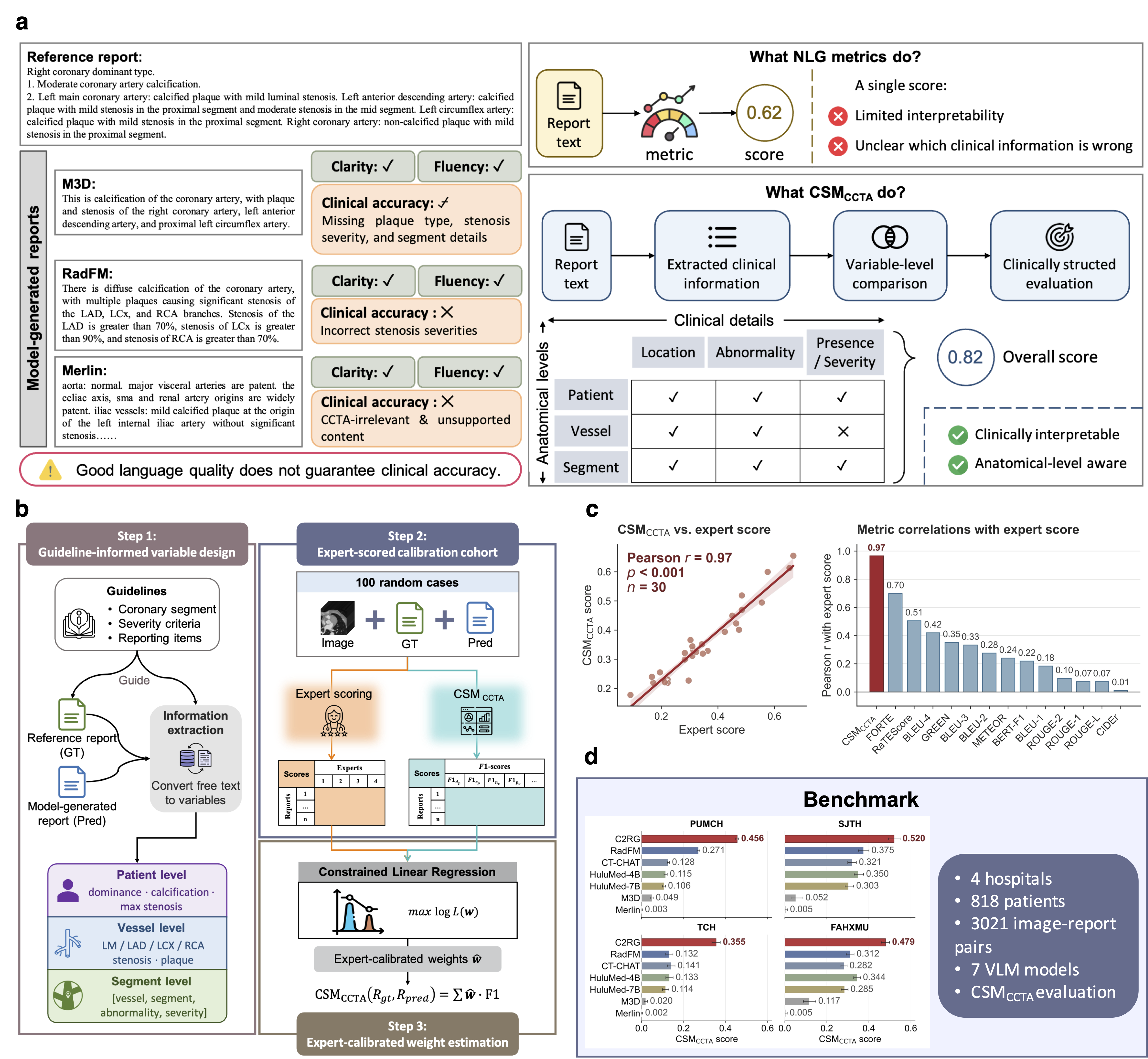}
\caption{\textbf{Motivation and overview of $\text{CSM}_{\text{CCTA}}$ and the multicenter benchmark.}
(a) Motivation and overview of CSM$_{\text{CCTA}}$ for clinically grounded evaluation of CCTA reports. Representative examples show that model-generated reports can remain clear and fluent despite substantial clinical errors. Conventional NLG metrics assess the entire report into a single score, while CSM$_{\text{CCTA}}$ extracts structured clinical information and performs variable-level comparisons across patient, vessel, and segment anatomical levels, enabling a more clinically interpretable assessment of report accuracy.
(b) Construction of $\text{CSM}_{\text{CCTA}}$: The proposed metric is developed in three steps. Step 1 uses guideline-informed variable design and clinical keyword extraction to parse free text into hierarchical clinical variables (patient-level summaries, vessel-level descriptions, and segment-level quadruples comprising vessel, segment, abnormality, and severity). Step 2 divides 100 expert-scored cases into a 70-case derivation set and a non-overlapping 30-case validation set. Step 3 estimates non-negative expert-calibrated weights by constrained linear regression using only the derivation set.
(c) On the held-out validation set, $\text{CSM}_{\text{CCTA}}$ demonstrates a strong correlation with experts (Pearson $r=0.97$, $p<0.001$), outperforming 13 compared metrics.
(d) $\text{CSM}_{\text{CCTA}}$ is used to benchmark seven representative VLMs across a multicenter dataset.
}
\label{fig:mainfig}
\end{figure}

\section*{Introduction}
Coronary computed tomography angiography (CCTA) transforms complex 3D anatomical images into essential information for the early screening, diagnosis, and clinical decision-making of coronary artery disease \cite{choi2008coronary,douglas2015outcomes,adamson2019guiding,serruys2021coronary}. While artificial intelligence has advanced various CCTA analysis tasks, including calcium scoring \cite{van2020deep,eng2021automated,xu2022automatic,van2023automatic}, plaque classification \cite{zreik2018recurrent,paul2022evaluation,lin2022deep}, CAD-RADS prediction \cite{cury2022cad,muscogiuri2020performance,van2023automatic,corti2025automated}, and cardiovascular event forecasting \cite{lin2022deep,miller2024patient,kim2025predicting}, the ultimate synthesis of these findings into high-quality clinical reports remains a complex, time-intensive burden for radiologists. Automated medical report generation (MRG) holds significant promise for alleviating this workload and improving efficiency. However, the successful development and subsequent clinical translation of such automated systems are facilitated by the availability of reliable evaluation metrics and standardized benchmarks.

Driven by advancements in foundation models, vision-language models (VLMs) have achieved notable progress in MRG. Yet, the majority of dedicated evaluation benchmarks and well-developed models focus predominantly on 2D imaging modalities, particularly chest X-rays \cite{wang2023r2gengpt,tanida2023interactive,li2023unify,li2024organ}. While general-purpose VLMs have been explored for multimodal and multi-anatomical structures \cite{zhou2023transferring,wu2024maken,yang2024pclmed}, their performance remains insufficient for accurate CCTA report generation in the evaluated settings. Although research into 3D medical images is expanding across brain CT \cite{li2025towards}, pulmonary CTA \cite{zhong2025vision}, multi-organ CT \cite{hamamci2024developing,bai2024m3d,wu2025towards}, and preliminary CCTA studies \cite{ye2024c2rg}, the lack of a dedicated CCTA benchmark forces models to rely on inconsistent evaluation settings, complicating cross-study comparison and preclinical evaluation.

Medical report evaluation requires metrics to prioritize clinical correctness over textual fluency. Large language models (LLMs) now generate reports with excellent clarity and fluency. Consequently, readability alone is insufficient for evaluating medical reports, making clinical accuracy a central consideration. In practice, models may produce highly readable text that contains severe clinical errors (Fig.~\ref{fig:mainfig}a). Therefore, the evaluation bottleneck lies in clinical accuracy. This requires the precise identification of abnormality types, anatomical locations, and severity gradings. 
Traditional natural language generation (NLG) metrics (e.g., BLEU \cite{papineni2002bleu}, ROUGE \cite{lin2004rouge}, METEOR \cite{banerjee2005meteor}, CIDEr \cite{vedantam2015cider}, and BERTScore (BERT-F1) \cite{zhang2019bertscore}) primarily focus on semantic similarity and fluency, but they do not explicitly account for clinically critical information. 
Furthermore, the medical-domain metrics including dataset-specific metrics (e.g., RadGraph and RadCliQ \cite{yu2023evaluating} for MIMIC-CXR \cite{johnson2019mimic}) and general-purpose metrics (e.g., RaTEScore \cite{zhao2024ratescore} and GREEN \cite{ostmeier2024green}), do not adequately capture the hierarchical and highly localized anatomy of coronary arteries. 
In clinical practice, a generated CCTA report may identify the correct abnormality but assign it to the wrong anatomical location or misstate its severity. These cases require evaluation of the vessel, segment, abnormality, and associated attributes of each finding. Reports may also differ in anatomical granularity, ranging from patient-level summaries to vessel- and segment-level descriptions. Existing metrics typically return a single similarity score and may therefore provide unreliable or poorly interpretable assessments when reports differ in reporting detail (Fig.~\ref{fig:mainfig}a).

\begin{table*}[!b]
\centering
\caption{\textbf{Summary of the CCTA report generation benchmark dataset.}}
\label{tab:dataset_summary}
\footnotesize
\setlength{\tabcolsep}{1.2pt}
\renewcommand{\arraystretch}{1.2}
\begin{tabularx}{\textwidth}{>{\raggedright\arraybackslash}m{0.18\textwidth}*{5}{>{\centering\arraybackslash}X}}
\toprule
\multirow{2}{*}{\textbf{Statistic}} & \multicolumn{4}{c}{\textbf{Dataset}} & \multirow{2}{*}{\textbf{Total}} \\
\cmidrule(l){2-5}
& \textbf{PUMCH} & \textbf{TCH} & \textbf{SJTH} & \textbf{FAHXMU} & \\
\midrule
Total patient-report pairs & 567 & 105 & 49 & 97 & 818 \\
\midrule
Total CTA series & 2121 & 430 & 147 & 323 & 3021 \\
\midrule
CTA series per patient & 3.7 (1--11) & 4.1 (1--17) & 3.0 (1--11) & 3.3 (1--17) & 3.7 (1--17) \\
\midrule
Examination date & 2023-03-07 -- 2024-03-29 & 2014-02-24 -- 2014-12-03 & 2014-01-06 -- 2014-12-26 & 2016-01-04 -- 2016-10-17 & / \\
\midrule
\multicolumn{6}{l}{\textit{Anatomical level of reference report}} \\
Patient & 0 (0.0) & 0 (0.0) & 0 (0.0) & 0 (0.0) & 0 (0.0) \\
Vessel & 130 (22.9) & 99 (94.3) & 2 (4.1) & 1 (1.0) & 232 (28.4) \\
Segment & 437 (77.1) & 6 (5.7) & 47 (95.9) & 96 (99.0) & 586 (71.6) \\
\midrule
\multicolumn{6}{l}{\textit{Coronary dominance}} \\
Right & 524 (92.4) & 95 (90.5) & 35 (71.4) & 0 (0.0) & 654 (80.0) \\
Left & 36 (6.3) & 5 (4.8) & 2 (4.1) & 0 (0.0) & 43 (5.3) \\
Balanced & 7 (1.2) & 5 (4.8) & 3 (6.1) & 0 (0.0) & 15 (1.8) \\
Unclear & 0 (0.0) & 0 (0.0) & 9 (18.4) & 97 (100.0) & 106 (13.0) \\
\midrule
\multicolumn{6}{l}{\textit{Overall coronary calcification severity}} \\
None & 206 (36.3) & 0 (0.0) & 0 (0.0) & 0 (0.0) & 206 (25.2) \\
Mild & 134 (23.6) & 0 (0.0) & 0 (0.0) & 0 (0.0) & 134 (16.4) \\
Moderate & 116 (20.5) & 0 (0.0) & 0 (0.0) & 0 (0.0) & 116 (14.2) \\
Severe & 111 (19.6) & 0 (0.0) & 0 (0.0) & 0 (0.0) & 111 (13.6) \\
Unclear & 0 (0.0) & 105 (100.0) & 49 (100.0) & 97 (100.0) & 251 (30.7) \\
\midrule
\multicolumn{6}{l}{\textit{Patient-level maximal stenosis severity}} \\
None & 134 (23.6) & 49 (46.7) & 4 (8.2) & 0 (0.0) & 187 (22.9) \\
Mild & 213 (37.6) & 22 (21.0) & 27 (55.1) & 59 (60.8) & 321 (39.2) \\
Moderate & 116 (20.5) & 0 (0.0) & 8 (16.3) & 29 (29.9) & 153 (18.7) \\
Severe / occlusion & 104 (18.3) & 1 (1.0) & 8 (16.3) & 9 (9.3) & 122 (14.9) \\
Unclear & 0 (0.0) & 33 (31.4) & 2 (4.1) & 0 (0.0) & 35 (4.3) \\
\bottomrule
\end{tabularx}
\begin{flushleft}
\footnotesize
CTA series per patient denotes the mean number of CCTA image series available for each patient. Values are reported as \textit{mean (minimum--maximum)}.
For anatomical level of reference report and the three clinical characteristics, values in parentheses are percentages based on patient-report pairs. Overall coronary calcification severity was recorded only when explicitly stated in the report. Localized calcified plaques were not extrapolated to overall calcification severity.
\end{flushleft}
\end{table*}

To bridge this gap, we make two primary contributions in this study. First, we construct a multicenter benchmark to facilitate standardized, reproducible comparisons of VLM performance in CCTA report generation. Second, to evaluate CCTA clinical findings, we propose a novel \textit{C}linically \textit{S}tructed \textit{M}etric ($\text{CSM}_{\text{CCTA}}$). 
Within the $\text{CSM}_{\text{CCTA}}$ framework, the metric automatically extracts essential clinical terms based on a guideline-informed variable design and organizes them into precise vessel-segment-abnormality-severity quadruples, alongside broader patient- and vessel-level variables (Fig.~\ref{fig:mainfig}b). 
Furthermore, $\text{CSM}_{\text{CCTA}}$ employs an adaptive hierarchical scoring mechanism that dynamically evaluates reports at their finest shared anatomical level and aggregates findings from more specific anatomical levels when necessary. This level-matched comparison distinguishes clinical accuracy from the anatomical level of reporting, preventing reports from being unfairly penalized when evaluated against less detailed reference reports.
Finally, to ensure that the evaluation aligns with clinical guidelines and practice priorities, the metric's internal weighting scheme is rigorously calibrated via constrained linear regression using a dedicated derivation set of expert-scored cases. 
Ultimately, this comprehensive approach enables anatomically resolved evaluation that preserves clinically meaningful localization while accommodating reports written at different anatomical levels.

We establish baseline performances (Fig.~\ref{fig:mainfig}d) on our proposed benchmark by evaluating 7 state-of-the-art VLMs (\textit{i.e.}, C2RG~\cite{ye2024c2rg}, CT-CHAT~\cite{hamamci2024developing}, M3D~\cite{bai2024m3d}, RadFM~\cite{wu2025towards}, HuluMed-4B/7B~\cite{jiang2025hulu}, and Merlin~\cite{blankemeier2026merlin}). The proposed $\text{CSM}_{\text{CCTA}}$ is then validated through extensive analyses and controlled perturbation experiments. Specifically, our results on a held-out validation set show that $\text{CSM}_{\text{CCTA}}$ maintains a strong correlation with human experts and achieved the highest observed correlation among the 13 compared metrics (Fig.~\ref{fig:mainfig}c).
Ultimately, this work provides a standardized evaluation setting that supports the continued development of clinically viable VLMs for CCTA report generation. Beyond this specific modality, the dynamic hierarchical evaluation strategy introduced by $\text{CSM}_{\text{CCTA}}$ offers a design template that may be adapted to other medical domains after task-specific schema design and expert recalibration.

\section*{Results}

\subsection*{Multicenter Benchmark for CCTA Report Generation}
To evaluate CCTA report generation, we established a multicenter benchmark comprising 3,021 CCTA series and 818 patient-report pairs from four hospitals (Table~\ref{tab:dataset_summary}). On this benchmark, we evaluated seven representative 3D medical VLMs (Fig.~\ref{fig:mainfig}d). The generated reports were assessed using our proposed $\text{CSM}_{\text{CCTA}}$ as the primary metric, alongside 13 established baselines (\textit{i.e.}, BLEU-1--4~\cite{papineni2002bleu}, ROUGE-1, ROUGE-2, ROUGE-L~\cite{lin2004rouge}, METEOR~\cite{banerjee2005meteor}, CIDEr~\cite{vedantam2015cider}, BERT-F1~\cite{zhang2019bertscore}, GREEN~\cite{ostmeier2024green}, RaTEScore~\cite{zhao2024ratescore}, and FORTE~\cite{zhong2025vision}). Specifically, $\text{CSM}_{\text{CCTA}}$ represents reports as hierarchical clinical variables and evaluates them at the finest shared anatomical level. It reports the anatomical evaluation level, and derives clinical accuracy by combining variable-level scores via expert-calibrated weights. Further details regarding benchmark construction and metric definitions are provided in \hyperref[sec:com_methods]{\textit{Methods}}.

\begin{figure*}[!htbp]
    \centering
    \includegraphics[width=\linewidth]{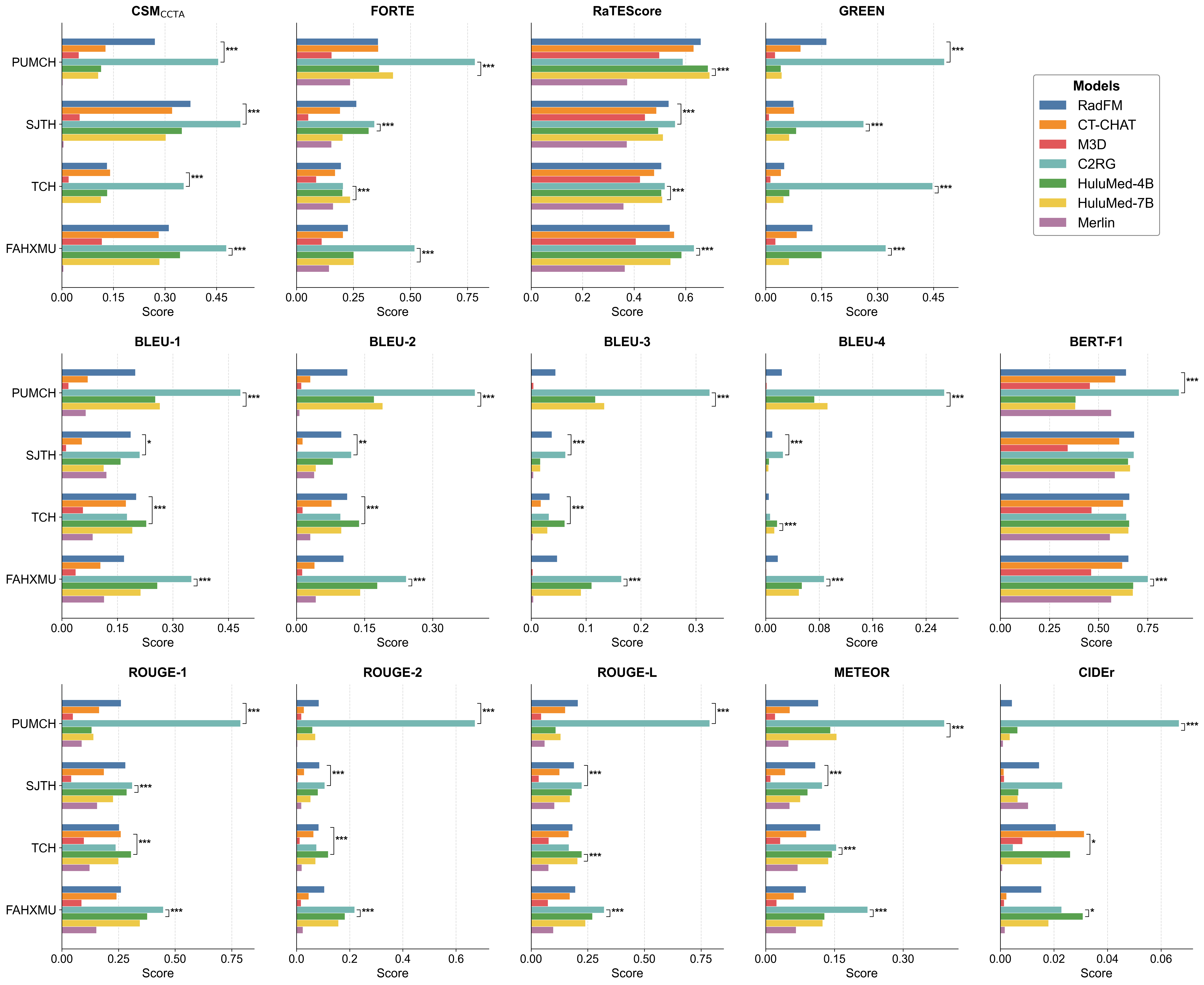}
    \caption{\textbf{Performance of 3D medical VLMs on the benchmark.} 
    Each panel summarizes one evaluation metric. Compared with NLG metrics, CSM$_{\text{CCTA}}$ provides more discriminative separation among models across these datasets. Significance brackets compare the highest- and second-highest-scoring models within each dataset, and asterisks indicate statistical significance ($*P<0.05$, $**P<0.01$, and $***P<0.001$).}
    \label{fig:model_dataset_metric_comparison}
\end{figure*}

\subsection*{Performance of 3D Medical VLMs}
To evaluate the performance of 3D medical VLMs within this benchmark, we compared CSM$_{\text{CCTA}}$ against existing metrics in assessing the quality and clinical relevance of generated CCTA reports. The results are summarized in Fig.~\ref{fig:model_dataset_metric_comparison} and Supplementary Table~\ref{supp-tab:comparison_results}.

\paragraph{Model comparison across the benchmark.}
On our benchmark, the 3D medical VLMs exhibited distinct performance disparities across the four datasets. Specifically, the domain-adapted C2RG model consistently achieved the highest CSM$_{\text{CCTA}}$ scores (PUMCH: $0.456$; SJTH: $0.520$; TCH: $0.355$; FAHXMU: $0.479$). In contrast, models primarily trained on out-of-domain data (\textit{e.g.}, Merlin, trained on abdominal images) generated limited relevant findings and performed poorly. Notably, compared metrics struggled to reflect this clear performance gap: semantic metrics (RaTEScore, BERT-F1) showed compressed, indiscriminately high score ranges, while lexical metrics (BLEU, ROUGE, METEOR, CIDEr) were overly sensitive to surface-level wording. This confirms that high textual similarity in traditional metrics does not equate to the clinical completeness of model-generated reports.

\paragraph{Analysis across anatomical levels with CSM$_{\text{CCTA}}$.}
To provide an anatomically related analysis of VLM performance, we stratified the report pairs by anatomical level (patient, vessel, or segment), as illustrated in Fig.~\ref{fig:com_statistical_analysis} and detailed in Supplementary Fig.~\ref{supp-fig:granularity_flow}.
Top-performing models like C2RG excel by producing highly detailed, fine-grained findings. Conversely, generalist models struggle with clinical depth. For instance, M3D yields only a minor subset of reports with superficial, patient-level information, providing limited CCTA-specific clinical information. Moreover, most outputs from M3D and Merlin are entirely irrelevant to CCTA, rendering them unevaluable (scoring 0). Ultimately, these results underscore that detailed finding descriptions are important for the evaluated CCTA reporting task, rather than merely generating fluent or superficial text.

\begin{figure*}[!t]
    \centering
    \includegraphics[width=\linewidth]{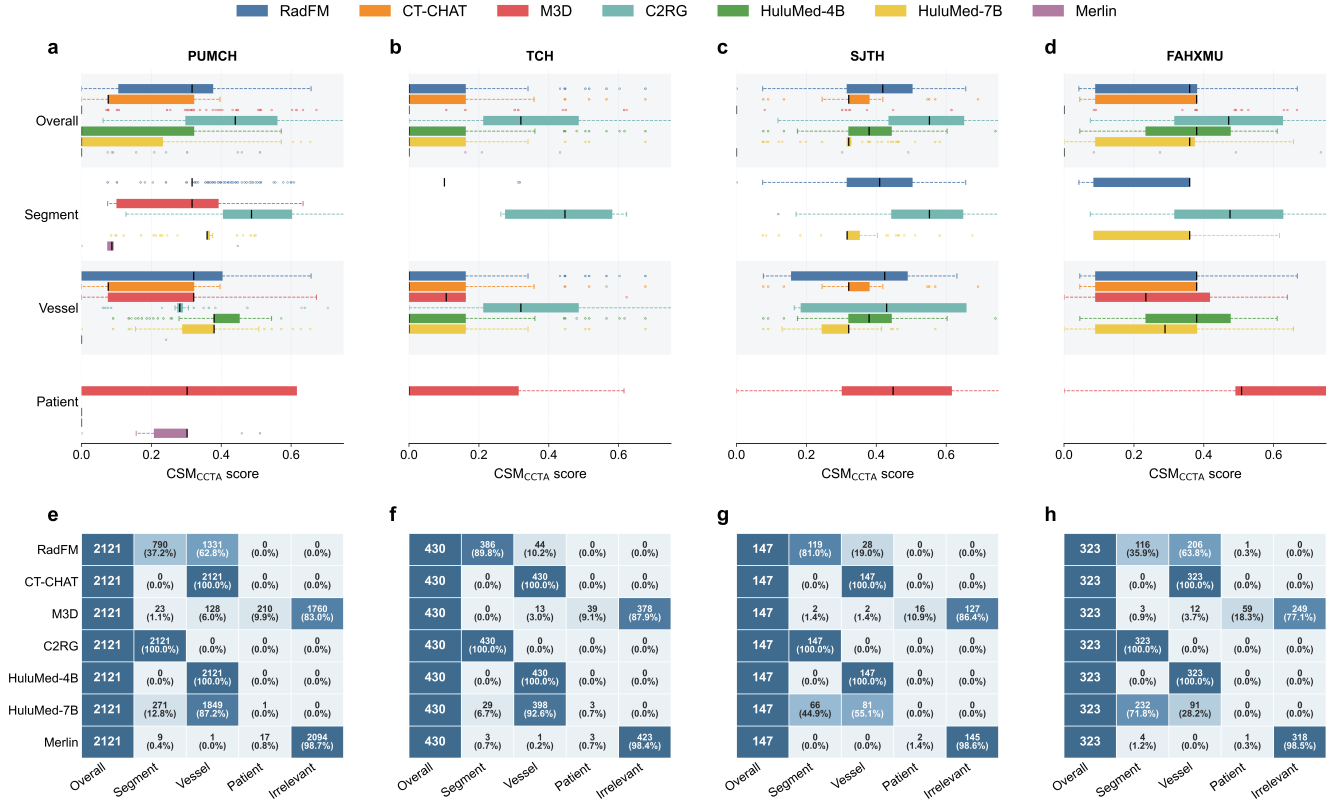}
    \caption{\textbf{CSM$_{\text{CCTA}}$ score distributions by anatomical level across datasets.}
    The upper panels display the CSM$_{\text{CCTA}}$ score distributions both overall and by evaluation level (patient, vessel, or segment). The lower panels show the number of report pairs evaluated at each level for each model. The results are grouped by four hospitals: \textbf{(a, e)} PUMCH, \textbf{(b, f)} TCH, \textbf{(c, g)} SJTH, and \textbf{(d, h)} FAHXMU. ``Irrelevant'' denotes report pairs for which no valid anatomical evaluation level could be established, and these pairs were assigned a CSM$_{\text{CCTA}}$ score of 0.
    } 
    \label{fig:com_statistical_analysis}
\end{figure*}

\subsection*{Clinical Relevance of Metrics}
\label{sec:ClinicalRelevance}
To determine whether the evaluation metrics reflect clinical concerns and align with expert judgments, we assembled an expert-scored cohort of 100 cases from PUMCH. Each case contained CCTA images, the reference report, and a C2RG-generated report. We prespecified 70 cases as a derivation set for estimating the CSM$_{\text{CCTA}}$ weights and reserved the remaining 30 cases as a non-overlapping validation set for correlation analysis. Report-level preference testing provided a complementary comparison of metric and expert decisions. Details of the scoring protocol, cohort allocation, and inter-expert reliability are provided in the Supplementary Materials.

\begin{figure}[!htbp]
    \centering
    \includegraphics[width=\linewidth]{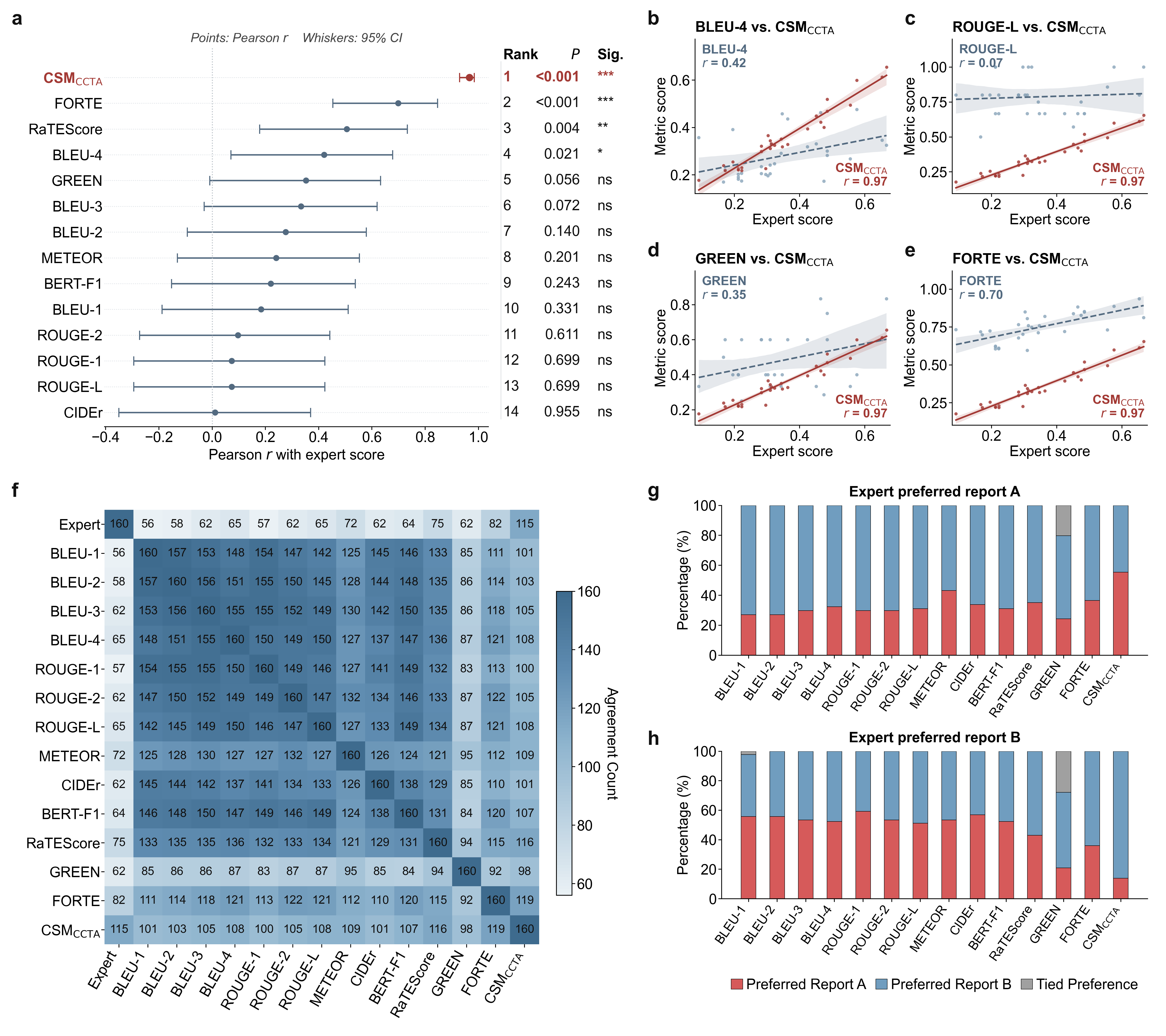}
    \caption{\textbf{Clinical relevance analysis among metrics against human experts.} 
    \textbf{(a)} Forest plot comparing the Pearson correlation coefficients ($r$) between expert scores and each evaluation metric. Points indicate the Pearson $r$ values, and whiskers represent the 95\% CI. The accompanying table details the performance rank, $P$ values, and statistical significance (Sig.) for each metric. 
    CSM$_{\text{CCTA}}$ (highlighted in red) demonstrates the highest observed correlation with expert clinical judgment, achieving the highest correlation and the narrowest CI.
    \textbf{(b-e)} Scatter plots providing sample-wise comparisons between expert scores and representative compared metrics: BLEU-4 (b), ROUGE-L (c), GREEN (d), and FORTE (e). Each panel contrasts a specific compared metric (blue) with CSM$_{\text{CCTA}}$ (red). The lines denote the linear regression fits for the respective metrics, with shaded regions representing the 95\% CI.
    \textbf{(f)} Pairwise agreement counts among expert preferences and automated metric preferences across 160 comparisons. $\text{CSM}_{\text{CCTA}}$ achieved the highest agreement with experts' clinical report-level preferences.
    \textbf{(g, h)} Metric-derived preferences stratified by cases in which experts preferred Report A (g) or Report B (h). Bars show the percentages of metric decisions favoring Report A (red), favoring Report B (blue), or indicating a tied preference (gray).}
    \label{fig:metric_correlation}
\end{figure}

On the held-out 30-case validation set, we calculated Pearson correlation coefficients between each metric and the average expert scores. As shown in Fig.~\ref{fig:metric_correlation}a, CSM$_{\text{CCTA}}$ achieved the highest observed correlation ($r = 0.97$, $p < 0.001$) and had the narrowest observed 95\% confidence interval (CI), substantially outperforming both traditional NLG (\textit{e.g.}, BLEU) and medical-specific metrics (\textit{e.g.}, FORTE).

To assess metric sensitivity to subtle clinical differences, both evaluation metrics and experts scored 160 randomly matched pairs of model-generated reports to identify the superior candidate.
As shown in Fig.~\ref{fig:metric_correlation}f, CSM$_{\text{CCTA}}$ achieved the highest agreement with human decisions (71.9\%, 115/160 pairs), consistently outperforming other metrics across all preference scenarios (Fig.~\ref{fig:metric_correlation}g\&h). For comparison, the strongest medical baseline, FORTE, aligned with experts in only 51.3\% of cases, while NLG metrics performed even worse. These results suggest that CSM$_{\text{CCTA}}$ better reflected expert preferences than the compared metrics in this pairwise evaluation.

Notably, the 71.9\% preference agreement rate does not fully mirror CSM$_{\text{CCTA}}$'s near-perfect global correlation. This discrepancy may partly reflect the metric’s greater weighting of vessel-level findings, which penalizes vessel-level errors significantly more heavily than segment-level inaccuracies. Consequently, when two candidate reports are identical at the vessel level but differ only in fine-grained segment descriptions, CSM$_{\text{CCTA}}$ becomes less sensitive than human experts, occasionally leading to diverging selections.

\begin{figure}[!t]
    \centering
    \includegraphics[width=\linewidth]{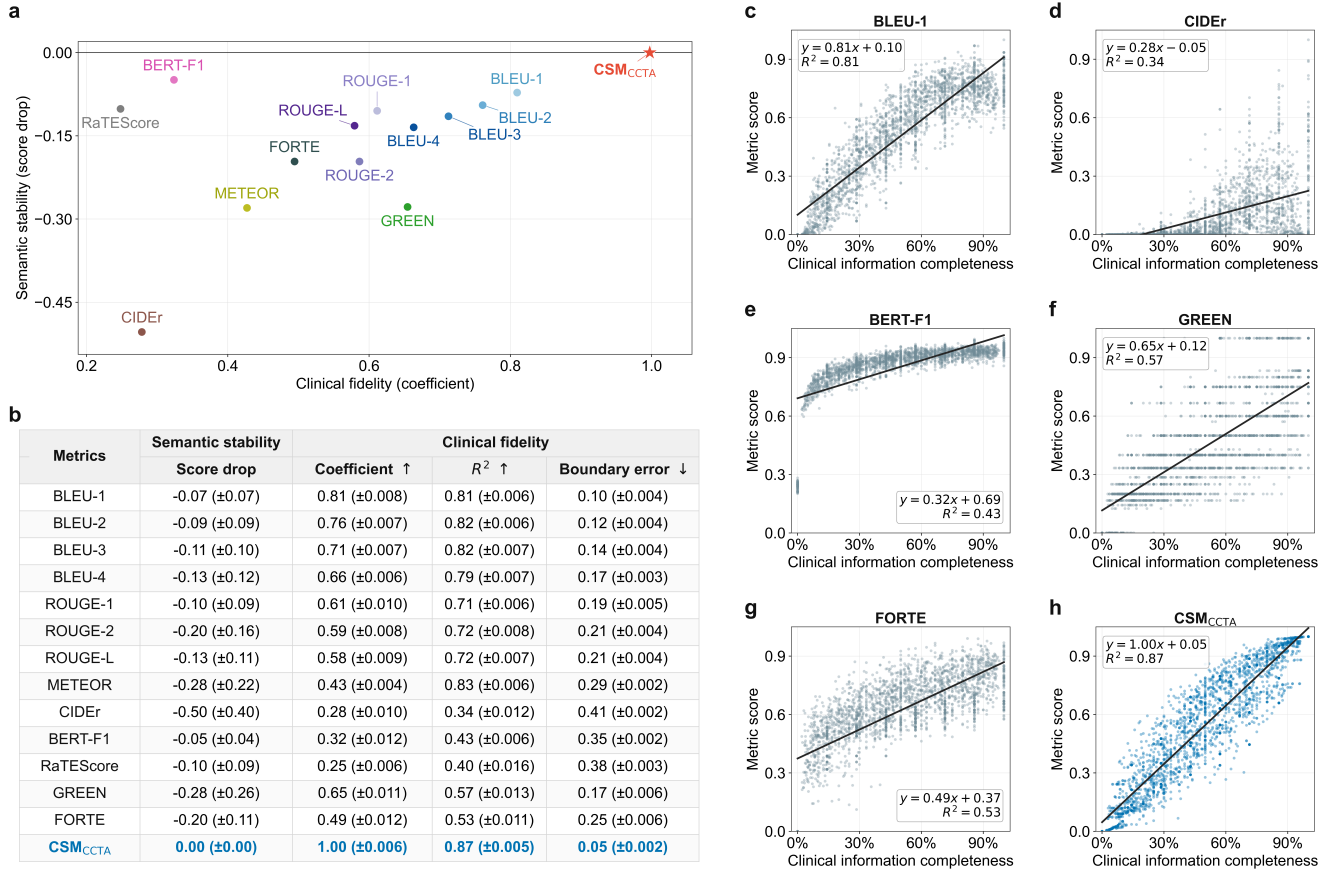}
    \caption{\textbf{Robustness analysis of metrics.} 
    \textbf{(a)} The scatter plot illustrates the holistic performance of $\text{CSM}_{\text{CCTA}}$ and compared metrics. The x-axis represents clinical fidelity (quantified by the linear regression coefficient), and the y-axis indicates semantic stability (measured by the score drop under semantic-preserving lexical variation). $\text{CSM}_{\text{CCTA}}$ (red star) demonstrates superior robustness by achieving optimal performance in both dimensions.
    \textbf{(b)} Quantitative comparison detailing the robustness properties across all metrics. Semantic stability is measured by the score drop following semantic-preserving lexical substitutions. Clinical fidelity is evaluated via linear regression analysis, characterized by three derived indicators: the linear regression coefficient (sensitivity to information variation), the $R^2$ value (reliability of the linear fit), and the boundary error (deviation from extreme theoretical limits at 0\% and 100\% information completeness).
    \textbf{(c--h)} Scatter plots detailing the clinical fidelity of representative compared metrics: BLEU-1 (c), CIDEr (d), BERT-F1 (e), GREEN (f), and FORTE (g), compared against $\text{CSM}_{\text{CCTA}}$ (h) under progressive clinical information omission. The x-axis denotes the completeness ratio of clinical information, and the y-axis shows the corresponding metric score. Black lines indicate the linear regression fits alongside their respective equations and $R^2$ values.}
    \label{fig:robustness_analysis}
\end{figure}

\subsection*{Robustness Analysis of Evaluation Metrics}
A reliable metric should remain invariant to semantic-preserving paraphrasing while being highly sensitive to the omission of critical clinical facts. By stress-testing the metrics on data from PUMCH, we demonstrate the robustness from two perspectives: semantic stability and clinical fidelity. Fig.~\ref{fig:robustness_analysis}a illustrates the holistic performance of CSM$_{\text{CCTA}}$ and 13 compared metrics.

\paragraph{Semantic stability.} 
In clinical practice, due to variations in radiologists' personal habits or institutional protocols, reports often employ diverse expressions to convey identical clinical findings. For example, qualitative descriptions like ``severe stenosis'' are occasionally detailed with quantitative values, such as ``80\% stenosis''. Semantic stability is the ability of a metric to remain invariant under such clinically equivalent lexical variations.
To quantify semantic stability, we constructed a modified test set where descriptive clinical keywords (\textit{i.e.}, degree adverbs) were replaced by equivalent numerical percentages at a 50\% replacement rate. Since the clinical semantics are preserved, the score drop after perturbation should ideally be zero.
As shown in Fig.~\ref{fig:robustness_analysis}a and \ref{fig:robustness_analysis}b, most compared metrics exhibit semantic rigidity, penalizing these clinically equivalent variations (e.g., the score of CIDEr drops by 0.50). In contrast, CSM$_{\text{CCTA}}$ demonstrates semantic stability with a score drop of 0.00.

\paragraph{Clinical fidelity.}
The ability of a metric to reflect the presence of critical clinical information is referred to as clinical fidelity, serving as an indicator of its robustness. We evaluated this property via a progressive information omission experiment, generating simulated reports by randomly removing key clinical information at proportions ranging from 0\% to 100\%. Clinical fidelity is evaluated via linear regression analysis utilizing three derived metrics shown in Fig.~\ref{fig:robustness_analysis}b. These encompass the linear regression coefficient ($\beta$) representing the sensitivity gradient to information loss, the coefficient of determination ($R^2$) indicating the reliability of the linear fit, and the boundary error ($E_{bound}$) measuring the deviation from the ideal theoretical bounds at 0\% and 100\% completeness. Fig.~\ref{fig:robustness_analysis}c-h illustrates the response patterns of representative metrics. Several baselines exhibit poor robustness. BERT-F1 demonstrates a largely unresponsive scoring pattern ($\beta = 0.32$, $E_{bound} = 0.35$) and fails to adequately distinguish between reports completely devoid of findings and perfectly accurate ones. Meanwhile, other metrics such as CIDEr and GREEN display highly scattered distributions with low $R^2$ values. Conversely, CSM$_{\text{CCTA}}$ scores maintain a stable linear alignment with clinical information completeness ($\beta = 1.00$), featuring a reliable linear fit ($R^2 = 0.87$) and the lowest boundary error ($0.05$) among the compared metrics under the tested omission procedure.

\subsection*{Illustrative Case Analysis of CSM$_{\text{CCTA}}$}

\begin{figure}[!b]
\centering
\includegraphics[width=\textwidth]{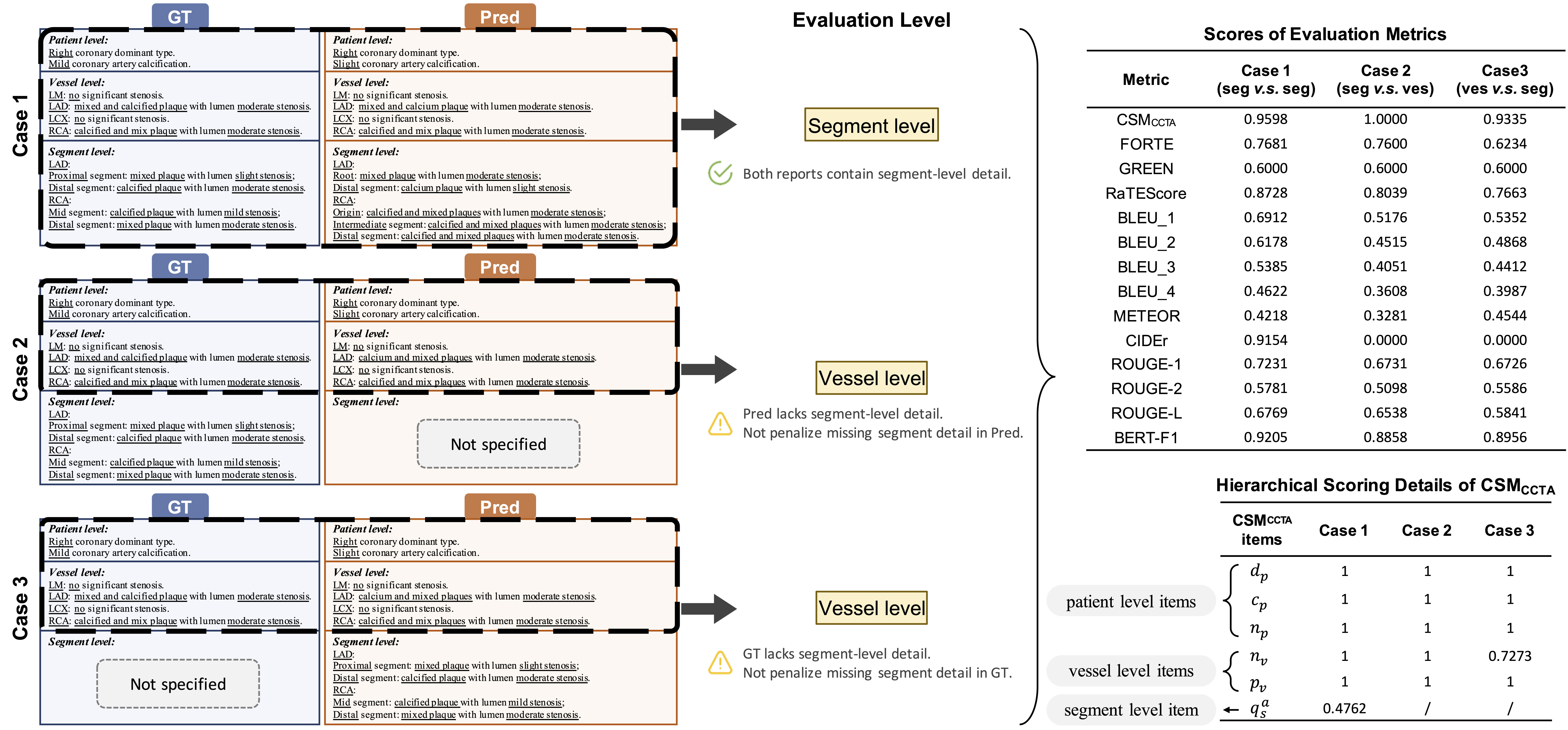}
\caption{\textbf{Application of the proposed CSM$_{\text{CCTA}}$ on representative coronary CTA reports.} The left panel displays three distinct case scenarios comparing Ground Truth (GT) and Predicted (Pred) reports, highlighting variations in the presence of fine-grained details. The middle section illustrates the dynamic strategy, determining the evaluation level based on the available clinical information to avoid unfair penalization for missing details. The top-right table compares the scores of various evaluation metrics across these three cases. The bottom-right table presents a detailed hierarchy of the CSM$_{\text{CCTA}}$ scores into patient-level, vessel-level, and segment-level items.}
\label{fig:application_case}
\end{figure}

To illustrate how CSM${\text{CCTA}}$ behaves when paired reports differ in anatomical specificity, we examined three report pairs (Fig.~\ref{fig:application_case}). In these examples, CSM${\text{CCTA}}$ selects the finest anatomical level shared by both reports, thereby reducing the influence of mismatched reporting detail on the resulting comparison.
In contrast, other metrics may not distinguish true clinical disagreement from differences in reporting detail. This limitation is illustrated by cases in which the reference and generated reports are expressed at different anatomical levels. For example, CIDEr drops from 0.9154 in Case 1 to 0 in Cases 2 and 3, while BLEU-4 decreases from 0.4622 to 0.3608 and 0.3987. Meanwhile, GREEN outputs an identical score of 0.6000 across all three pairs. Unlike these metrics, CSM$_{\text{CCTA}}$ maintains valid clinical comparisons while explicitly reporting the evaluated anatomical level, providing a more interpretable evaluation when reference and generated reports differ in detail.

In addition to paired-report comparison, CSM$_{\text{CCTA}}$ can also operate in a single-report mode. In this setting, no reference report is provided and clinical accuracy is therefore not assessed. Instead, CSM$_{\text{CCTA}}$ characterizes the presence of clinical variables and the anatomical specificity of the report. This functionality enables researchers to determine whether an individual report meets the anatomical specificity required by a predefined task.

\section*{Discussion}
This study addresses two related barriers to evaluating CCTA report generation. The multicenter benchmark provides a common setting for comparing models across institutions. CSM$_{\text{CCTA}}$ complements this resource by converting free-text reports into patient-, vessel-, and segment-level clinical variables. Its central contribution is an adaptive comparison at the finest anatomical level supported by both reports. This design separates the accuracy of reported findings from the level of anatomical detail at which they are expressed.

The benchmark results illustrate why this distinction matters for model development. The CCTA-adapted model generated more relevant and anatomically specific findings, whereas several generalist models produced coarse or unrelated reports. This pattern suggests that fluent report-style language alone is an inadequate indicator of report quality in a specialized imaging task. Anatomical-level outputs provide additional information beyond a scalar score by revealing whether a model fails in overall relevance, vessel-level description, or segment-level localization. These failure modes may provide more actionable diagnostic information than lexical similarity alone when selecting data, training objectives, or models for further development. However, the observed ranking applies to the seven evaluated models and should not be interpreted as a universal comparison of model architectures.

The expert analyses provide held-out internal validation evidence for the clinical relevance of CSM$_{\text{CCTA}}$, with scores derived from the calibration cohort remaining strongly aligned with expert assessments in the non-overlapping validation cohort and better reflecting expert report preferences than existing metrics. The lower pairwise agreement, however, suggests that the current weighting scheme may be less sensitive to subtle segment-level differences due to the greater emphasis on vessel-level findings. The perturbation and case analyses further show that CSM$_{\text{CCTA}}$ is robust to clinically equivalent wording, sensitive to omitted findings, and avoids penalizing differences in reporting detail through adaptive level matching. This flexibility also introduces an interpretive trade-off, as a report may be accurate at the shared anatomical level while lacking the detail required for a specific task. CSM$_{\text{CCTA}}$ should therefore be interpreted together with its selected anatomical level and component outputs. In single-report mode, it can assess structural completeness and anatomical specificity, but not clinical accuracy in the absence of a reference report.

Several limitations constrain the scope of these conclusions. CSM$_{\text{CCTA}}$ relies on a CCTA-specific schema, and adaptation to other imaging examinations will require task-specific definitions, extraction rules, and expert recalibration. Deterministic clinical keyword extraction is transparent and reproducible, but it may miss uncommon, ambiguous, or institution-specific expressions. The expert derivation and validation cohorts, as well as the perturbation experiments, were drawn from PUMCH. Larger multi-institutional expert studies are needed to assess the transportability of the learned weights. Reference reports also vary in completeness and anatomical detail, so level matching reduces but does not eliminate dependence on the reference standard. Finally, the real-world examples were illustrative rather than a prospective evaluation of clinical utility.

Taken together, the benchmark and CSM$_{\text{CCTA}}$ provide a clinically structured basis for comparing CCTA report-generation models. Their principal value lies in making both finding agreement and reporting specificity visible, while preserving clear boundaries on what each evaluation mode can establish.

\section*{Methods}
\label{sec:methds}

\subsection*{Construction of the Multicenter CCTA Report Generation Benchmark}
\phantomsection
\label{sec:benchmark_methods}
The benchmark is constructed for evaluating CCTA report generation. Each model received a 3D CCTA image as input and generated a radiology report for comparison with the corresponding clinical reference report.

\paragraph{Dataset.}
After excluding cases with inadequate image quality, we retained all eligible cases from Peking Union Medical College Hospital (PUMCH), Tianjin Chest Hospital (TCH), Beijing Shijitan Hospital (SJTH), and The First Affiliated Hospital of Xinjiang Medical University (FAHXMU). 
A subset of 100 cases was randomly sampled from the PUMCH dataset to serve as the expert evaluation cohort, completely separate from the benchmark (Supplementary Fig.~\ref{supp-fig:data}). The final benchmark comprised 3,021 CCTA series from 818 patient-report pairs across four hospitals. The cohorts included PUMCH ($n=567$), TCH ($n=105$), SJTH ($n=49$), and FAHXMU ($n=97$). 
Table~\ref{tab:dataset_summary} summarizes the scan statistics and report-derived coronary findings.

\paragraph{Models.}
We evaluated seven open-source 3D medical VLMs, covering both CCTA-adapted and generalist models.
C2RG~\cite{ye2024c2rg} is a 3D VLM specifically adapted for CCTA report generation.
CT-CHAT~\cite{hamamci2024developing} is a conversational VLM for 3D chest CT that supports interactive visual question answering (VQA) and report-style reasoning.
M3D~\cite{bai2024m3d} is a unified multimodal large language model capable of diverse 3D tasks, including retrieval, report generation, VQA, localization, and segmentation.
RadFM~\cite{wu2025towards} is a generalist radiology foundation model designed to extract clinically oriented representations for broad downstream radiology tasks.
HuluMed-4B/7B~\cite{jiang2025hulu} are generalist medical VLMs that integrate 2D, 3D, and video inputs for diverse medical reasoning tasks.
Merlin~\cite{blankemeier2026merlin} is a 3D VLM for computed tomography that supports adapted and zero-shot tasks, including segmentation and report generation.

\paragraph{Evaluation metrics.}
Generated reports were compared with their paired clinical reference reports using CSM$_{\text{CCTA}}$ as the primary metric alongside other 13 metrics. 
These metrics include 10 NLG metrics and 3 medical-specific metrics. Among the NLG metrics, BLEU-1, BLEU-2, BLEU-3, BLEU-4~\cite{papineni2002bleu}, ROUGE-1, ROUGE-2, and ROUGE-L~\cite{lin2004rouge}, METEOR~\cite{banerjee2005meteor}, and CIDEr~\cite{vedantam2015cider} capture complementary forms of lexical overlap, while BERT-F1~\cite{zhang2019bertscore} measured contextual token similarity. The three medical report metrics included GREEN~\cite{ostmeier2024green}, RaTEScore~\cite{zhao2024ratescore}, and FORTE~\cite{zhong2025vision}. Since FORTE was originally developed for pulmonary CTA reports using a task-specific keyword list, we adapted it for CCTA evaluation by replacing the keyword list while retaining the scoring procedure. This adaptation was intended to align the terminology coverage of FORTE with the coronary anatomy and clinical findings represented in our benchmark.

\subsection*{Clinically Structured Metric (CSM$_{\text{CCTA}}$)}
\label{sec:com_methods}
\begin{figure*}[!t]
\centering
\includegraphics[width=0.75\textwidth]{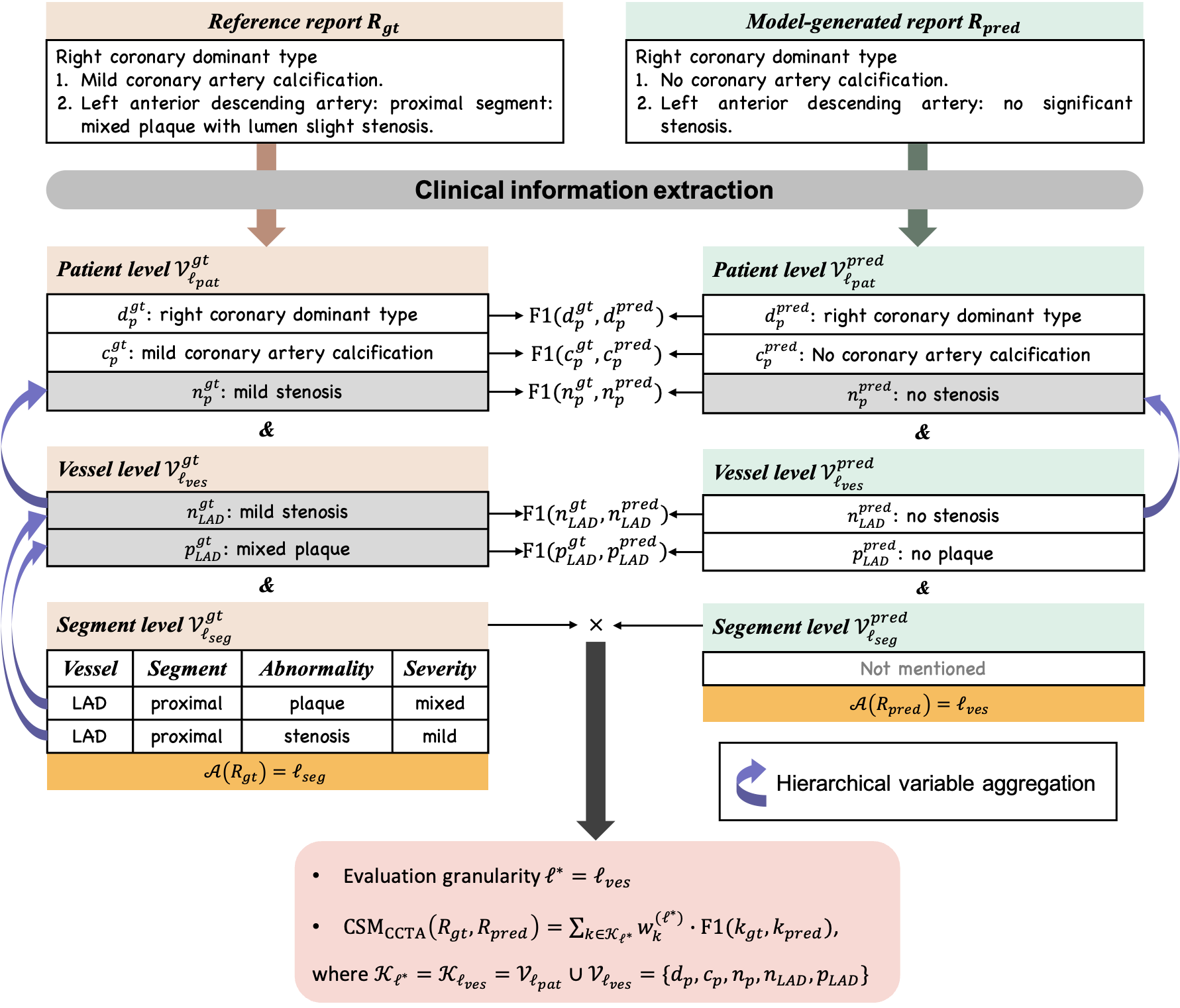}
\caption{\textbf{Workflow of CSM$_{\text{CCTA}}$.} Hierarchical clinical variables are extracted from the reference and model-generated reports to determine their anatomical levels and the shared evaluation level. Variable-level F1 scores are then combined using expert-calibrated weights to obtain the final CSM$_{\text{CCTA}}$ score.}
\label{fig:com_flow_workflow}
\end{figure*}

We propose CSM$_{\text{CCTA}}$, a hierarchical evaluation framework designed to assess CCTA report quality across three anatomical levels: patient, vessel, and segment levels. CSM$_{\text{CCTA}}$ incorporates bottom-up information aggregation and an adaptive scoring mechanism that quantifies agreement with the clinical findings represented in the reference report (Fig.~\ref{fig:com_flow_workflow}).

\subsubsection*{Hierarchical Variable Representation}
To evaluate a report $R$, we map it into a structured representation defined over a hierarchy of levels $\mathcal{L} = \{\ell_{pat}, \ell_{ves}, \ell_{seg}\}$, ordered by increasing anatomical specificity from the patient level to the vessel and segment levels. Let $\mathcal{V}_\ell$ denote the set of clinical variables at level $\ell \in \mathcal{L}$. 
The clinical information in $R$ is extracted into the following collection of variables, providing a structured representation:
\begin{align}
    \mathcal{V}_{\ell_{pat}} &= \{d_p, c_p, n_p\}, \\
    \mathcal{V}_{\ell_{ves}} &= \{n_v, p_v \mid v \in \mathbb{V}\}, \\
    \mathcal{V}_{\ell_{seg}} &= \bigcup_{v \in \mathbb{V}} \{q_s^a=(v, s, a, e_a) \mid s \in \mathbb{S}_v \},
\end{align}
where $\mathbb{V} = \{\text{LM}, \text{LAD}, \text{LCX}, \text{RCA}\}$ represents the main coronary vessels, and $\mathbb{S}$ represents the 18-segment coronary artery model\footnote{Right Coronary Artery (RCA): proximal (1), mid (2), distal (3), right Posterior Descending Artery (PDA) (4), right Posterolateral Branch (PLB) (16); Left Main (LM) (5); Left Anterior Descending (LAD): proximal (6), mid (7), distal (8), First Diagonal (D1) (9), Second Diagonal (D2) (10); Left Circumflex (LCx): proximal (11), First Obtuse Marginal (OM1) (12), distal (13), Second Obtuse Marginal (OM2) (14), left Posterior Descending Artery (PDA) (15), left Posterolateral Branch (PLB) (18); and Ramus Intermedius (RI) (17).} 
defined by the Society of Cardiovascular Computed Tomography (SCCT)~\cite{leipsic2014scct}. Here, $\mathbb{S}_v \subseteq \mathbb{S}$ denotes the SCCT segments anatomically assigned to vessel $v$. $\mathcal{V}_{\ell_{seg}}$ is composed of all extracted semantic quadruples across vessels. The specific clinical variables are defined as follows:

\begin{itemize}
    \item \textbf{Coronary Dominance ($d_p$):} $d_p \in \{\text{right}, \text{left}, \text{balanced}\}$.
    
    \item \textbf{Overall Coronary Calcification Severity ($c_p$):} The global ordinal severity of calcification for a patient, where $c_p \in \{0 : \text{none}, 1 : \text{mild}, 2 : \text{moderate}, 3 : \text{severe}\}$.
    
    \item \textbf{Maximal Stenosis Severity ($n_p, n_v$):} The highest ordinal grade of stenosis evaluated at both the patient level ($n_p$) and the individual vessel level ($n_v$), where $n_p, n_v \in \{0 : \text{none}, 1 : \text{mild}, 2 : \text{moderate}, 3 : \text{severe}\}$.
    
    \item \textbf{Plaque Type ($p_v$):} The multi-label plaque composition at the vessel level, where $p_v \subseteq \{\text{calcified}, \text{non-calcified}, \text{mixed}\}$.
    
    \item \textbf{Segment-level Semantic Quadruple ($q_s^a$):} Each quadruple is defined as $q_s^a = (v, s, a, e_a)$, where $v \in \mathbb{V}$ indicates the vessel and $s \in \mathbb{S}_v$ specifies the corresponding finer-level anatomical segment or branch under that vessel. The abnormality $a$ and its corresponding type or severity $e_a$ are defined via the following conditional mapping:
    \begin{align}
        a &\in \{\text{plaque}, \text{stenosis}\}, \\
        e_a &\in \begin{cases} 
            \mathcal{P}(\{\text{calcified}, \text{non-calcified}, \text{mixed}\}), & \text{if } a = \text{plaque}, \\ 
            \{0 : \text{none}, 1 : \text{mild}, 2 : \text{moderate}, 3 : \text{severe}\}, & \text{if } a = \text{stenosis}, 
        \end{cases}
    \end{align}
    where $\mathcal{P}(\cdot)$ denotes the power set (\textit{i.e.}, the set of all possible subsets), representing all potential combinations of plaque types.
\end{itemize}

\subsubsection*{Hierarchical Variable Aggregation}
CSM$_{\text{CCTA}}$ uses $\text{Agg}(\cdot)$ to complete missing coarser variables from reported segment level findings. Explicitly extracted variables are retained as the primary evidence. If a coarser variable is not explicitly reported, its value is inferred from the available segment level or vessel level variables. If a variable is neither reported nor derivable from more detailed findings, CSM$_{\text{CCTA}}$ leaves it uninstantiated. During scoring, an uninstantiated abnormality is counted as no abnormality when the variable is required for comparison.

The aggregation operator is defined only for coarser variables that can be derived from lower anatomical levels. For each vessel $v$, the segment quadruples in $\mathcal{V}_{\ell_{seg}}$ are separated into stenosis and plaque subsets:
\begin{align}
    \mathcal{Q}^{n}_v &= \{q_s^a=(v,s,a,e_a) \in \mathcal{V}_{\ell_{seg}} \mid s \in \mathbb{S}_v,\ a=\text{stenosis}\}, \\
    \mathcal{Q}^{p}_v &= \{q_s^a=(v,s,a,e_a) \in \mathcal{V}_{\ell_{seg}} \mid s \in \mathbb{S}_v,\ a=\text{plaque}\}.
\end{align}
Given these subsets, $\text{Agg}(\cdot)$ is specified for each completed variable. For vessel stenosis, CSM$_{\text{CCTA}}$ takes the maximal stenosis severity among the vessel segments. For patient stenosis, CSM$_{\text{CCTA}}$ takes the maximal completed vessel stenosis:
\begin{align}
    \text{Agg}_{n_v}(\mathcal{Q}^{n}_v) &=
    \begin{cases}
        \max\{e_a \mid q_s^a = (v,s,a,e_a) \in \mathcal{Q}^{n}_v\}, & \mathcal{Q}^{n}_v \neq \varnothing, \\
        \varnothing, & \mathcal{Q}^{n}_v = \varnothing,
    \end{cases} \\
    \text{Agg}_{n_p}(\{\tilde{n}_v\}_{v \in \mathbb{V}}) &=
    \begin{cases}
        \max\{\tilde{n}_v \mid v \in \mathbb{V}, \tilde{n}_v \neq \varnothing\}, & \exists v,\ \tilde{n}_v \neq \varnothing, \\
        \varnothing, & \text{otherwise}.
    \end{cases}
\end{align}
For vessel plaque, CSM$_{\text{CCTA}}$ takes the union of plaque types over the corresponding segment findings:
\begin{align}
    \text{Agg}_{p_v}(\mathcal{Q}^{p}_v) =
    \begin{cases}
        \bigcup\{e_a \mid q_s^a = (v,s,a,e_a) \in \mathcal{Q}^{p}_v\}, & \mathcal{Q}^{p}_v \neq \varnothing, \\
        \varnothing, & \mathcal{Q}^{p}_v = \varnothing.
    \end{cases}
\end{align}
The completed vessel and patient variables are then obtained by giving priority to explicitly extracted variables:
\begin{align}
    \tilde{n}_v &=
    \begin{cases}
        n_v, & \text{if } n_v \text{ is explicitly extracted from } R, \\
        \text{Agg}_{n_v}(\mathcal{Q}^{n}_v), & \text{otherwise},
    \end{cases} \\
    \tilde{p}_v &=
    \begin{cases}
        p_v, & \text{if } p_v \text{ is explicitly extracted from } R, \\
        \text{Agg}_{p_v}(\mathcal{Q}^{p}_v), & \text{otherwise},
    \end{cases} \\
    \tilde{n}_p &=
    \begin{cases}
        n_p, & \text{if } n_p \text{ is explicitly extracted from } R, \\
        \text{Agg}_{n_p}(\{\tilde{n}_v\}_{v \in \mathbb{V}}), & \text{otherwise}.
    \end{cases}
\end{align}
When comparison requires a stenosis label for an absent finding, $\varnothing$ is mapped to the normal category $0$.
This procedure preserves explicitly stated findings while allowing detailed segment descriptions to support the corresponding vessel and patient level variables. As a result, reports with different levels of anatomical detail can be evaluated within the same hierarchical representation (Supplementary Fig.~\ref{supp-fig:com_flow_aggregation}).

\subsubsection*{Adaptive Scoring across Anatomical Levels}
CSM$_{\text{CCTA}}$ first determines the finest anatomical level represented in the ground-truth and predicted reports.
Before selecting the anatomical evaluation level, CSM$_{\text{CCTA}}$ verifies that the report pair contains comparable CCTA-related clinical information. A report pair is labeled ``Irrelevant'' when no valid anatomical level can be established because the model-generated and reference reports contain no comparable CCTA-related clinical information.
Let $\mathcal{A}(R) \in \mathcal{L}$ denote the finest anatomical level at which report $R$ provides explicit clinical information. The anatomical levels represented in the reports are therefore $\mathcal{A}(R_{gt})$ and $\mathcal{A}(R_{pred})$. CSM$_{\text{CCTA}}$ evaluates the two reports at the coarser of these levels, defined as the selected anatomical evaluation level $\ell^*$:
\begin{eqnarray}
\ell^* = \min_{\prec} \{\mathcal{A}(R_{gt}), \mathcal{A}(R_{pred})\},
\end{eqnarray}
where $\ell_{pat} \prec \ell_{ves} \prec \ell_{seg}$. This rule avoids rewarding a prediction for segment level details when the ground truth is only reported at the vessel level, and it also avoids penalizing a prediction for missing segment level details when the prediction itself only supports a coarser comparison.

For each evaluation level, CSM$_{\text{CCTA}}$ uses the variables available up to that level. The variable set used for scoring is defined as
\begin{align}
    \mathcal{K}_{\ell_{pat}} &= \mathcal{V}_{\ell_{pat}}, \\
    \mathcal{K}_{\ell_{ves}} &= \mathcal{V}_{\ell_{pat}} \cup \mathcal{V}_{\ell_{ves}}, \\
    \mathcal{K}_{\ell_{seg}} &= \mathcal{V}_{\ell_{pat}} \cup \mathcal{V}_{\ell_{ves}} \cup \mathcal{V}_{\ell_{seg}}.
\end{align}
Thus, patient level scoring uses only patient variables, vessel level scoring uses patient and vessel variables, and segment level scoring uses patient, vessel, and segment variables.

Some variables have multiple anatomical instances. This multiplicity is handled when computing F1, without introducing additional variables. For vessel-level variables, all valid vessel-specific instances are evaluated jointly: $\text{F1}(n_v)$ is computed by pooling stenosis matches and errors across the four main vessels, and $\text{F1}(p_v)$ is computed analogously for plaque. For segment-level semantic quadruples, $\text{F1}(q_s^a)$ is computed over the pooled set of segment-level quadruple instances. If a variable has no valid instance for comparison, its F1 is treated as unavailable; the variable is excluded from the weighted sum, and the remaining weights are renormalized within the selected evaluation level. Further implementation details are provided in Supplementary Materials.

Each evaluation level has its own expert-calibrated weight vector over the corresponding variable set:
\begin{eqnarray}
\label{eq:weight}
\mathbf{w}^{(\ell)} = \{w_k^{(\ell)} \mid k \in \mathcal{K}_{\ell}\}, \quad
w_k^{(\ell)} \ge 0,\quad
\sum_{k \in \mathcal{K}_{\ell}} w_k^{(\ell)} = 1.
\end{eqnarray}
The estimation of these three weight vectors is described in the following \hyperref[sec:weight_estimation]{section}.

Let $\mathcal{I}_{\ell^*} \subseteq \mathcal{K}_{\ell^*}$ denote the variables with valid F1 scores in the current comparison. The weights are renormalized over these available variables:
\begin{eqnarray}
\bar{w}_k^{(\ell^*)}
=
\frac{w_k^{(\ell^*)}}{\sum_{j \in \mathcal{I}_{\ell^*}} w_j^{(\ell^*)}},
\quad k \in \mathcal{I}_{\ell^*}.
\end{eqnarray}
The final CSM$_{\text{CCTA}}$ score is then computed as:
\begin{eqnarray}
\text{CSM$_{\text{CCTA}}$}(R_{gt}, R_{pred}) =
\sum_{k \in \mathcal{I}_{\ell^*}} \bar{w}_k^{(\ell^*)} \cdot \text{F1}(k_{gt}, k_{pred}).
\end{eqnarray}
Along with this score, CSM$_{\text{CCTA}}$ records $\mathcal{A}(R_{gt})$ and $\mathcal{A}(R_{pred})$ to indicate the anatomical levels of the reference and generated reports.

\subsubsection*{Expert-Calibrated Weight Estimation}
\label{sec:weight_estimation}
To determine the importance of clinical variables across different evaluation levels as defined in Eq.~\ref{eq:weight}, we employed a data-driven calibration procedure to estimate the weights $w_{k}^{(\ell)}$ and align CSM$_{\text{CCTA}}$ with clinical judgment. This calibration used only the 70-case derivation set from the fine-grained expert scoring task. The non-overlapping 30-case validation set was reserved for the \hyperref[sec:ClinicalRelevance]{correlation analysis}. The expert scoring protocols are provided in Supplementary Materials.

Specifically, we used constrained linear regression without an intercept to model the relationship between the F1 scores of realized clinical variables and the overall scores assigned by experts. Both a non-negativity constraint ($w \ge 0$) and a zero-intercept constraint were imposed during optimization. The structured clinical variables were defined at the patient, vessel, and segment levels as described above, with the segment-level component computed from the pooled semantic quadruples $q_s^a$. We performed 5-fold cross-validation within the 70-case derivation set to assess the stability of the estimated weights at each level. The regression results, estimated weights, and level-wise interpretation are provided in Supplementary Materials.

\subsection*{Expert Scoring Protocol and Reliability Assessment}
Four radiologists evaluated the expert evaluation cohort using a six-question questionnaire that separately assessed plaque and stenosis localization and description accuracy. The complete allocation procedure, questionnaire design, and scoring criteria are provided in Supplementary Materials and Supplementary Table~\ref{supp-tab:questionnaire}. Reliability was assessed using ICC(2,1), Krippendorff's alpha, pairwise Cohen's kappa, and score-distribution analyses, with full results reported in Supplementary Fig.~\ref{supp-fig:expert_reliability}.

\bibliography{references}

\end{document}